\documentclass[twocolumn,10pt,a4paper]{article}

\usepackage[left=15mm, right=15mm, top=20mm, bottom=25mm, columnsep=18pt]{geometry}

\usepackage{helvet}             
\usepackage{mathptmx}           
\DeclareMathAlphabet{\mathcal}{OMS}{cmsy}{m}{n} 
\usepackage{microtype}          

\usepackage{cite}                         

\usepackage{amsmath,amssymb,amsfonts,amsthm}
\usepackage{bbm, mathtools}

\usepackage{graphicx}
\usepackage{subcaption}
\usepackage{booktabs}           
\usepackage{tabularx}

\usepackage{titlesec}
\titleformat{\section}{\large\sffamily\bfseries}{\thesection.}{0.5em}{}
\titleformat{\subsection}{\normalsize\sffamily\bfseries}{\thesubsection.}{0.5em}{}
\titlespacing*{\section}{0pt}{12pt}{6pt}
\titlespacing*{\subsection}{0pt}{10pt}{4pt}

\usepackage[labelfont={bf,sf}, labelsep=period, font=small, singlelinecheck=false]{caption}
\DeclareCaptionLabelFormat{naturelabel}{\sffamily\bfseries Fig.~#2}
\usepackage{xcolor}
\usepackage[
    colorlinks=true,
    allcolors=blue!80!black,
    linkcolor=blue,
    filecolor=magenta,
    urlcolor=cyan,
    pdftitle={Overleaf Example},
    pdfpagemode=FullScreen
]{hyperref}

\makeatletter
\renewcommand{\@maketitle}{%
  \vbox to 0pt{}%
  \vspace{-1em}%
  {\noindent\LARGE\sffamily\bfseries \@title \par}%
  \vskip 1em%
  {\noindent\normalsize\sffamily \@author \par}%
  \vskip 1.5em%
}
\makeatother

\title{Into My Arms: Structured reinforcement learning for autonomous transport of objects and humans}

\author{%
  {Elle Miller}$^{1*}$, 
  {Reem Almheiri}$^{2,3}$,
  {Takuma Tsukakoshi}$^{2}$,
  {Ben Sanati}$^{1}$,
  {Tamon Miyake}$^{2}$,
  {Yushi Wang}$^{2}$,
  {Namiko Saito}$^{2,4}$, 
  {Shigeki Sugano}$^{2}$,
  {Tetsuya Ogata}$^{2}$,
  {Rika Antonova}$^{5}$,
  {Oisin Mac Aodha}$^{1}$, \& 
  {Sethu Vijayakumar}$^{1}$\\[0.5em]
  \small $^{1}$University of Edinburgh, UK.\\
  \small $^{2}$Waseda University, Tokyo, Japan.\\
  \small $^{3}$Technology Innovation Institute, Abu Dhabi.\\
  \small $^{4}$Microsoft Research Asia, Tokyo, Japan.\\
  \small $^{5}$University of Cambridge, UK.\\
  \small $^{*}$e-mail: elle.miller@ed.ac.uk
}

\date{} 

\usepackage{CJKutf8}
\usepackage{xcolor}
\usepackage[colorinlistoftodos]{todonotes}
\definecolor{pinkcomment}{rgb}{1.0, 0.4, 0.7} 

\begin{document}

\twocolumn[
  \maketitle
  
  \noindent\textbf{Abstract.}
  \textbf{
  Whole-body humanoid manipulation remains bottlenecked by high-dimensional state-action spaces, contact-rich environments, and long task horizons. This trilemma is a challenge that state-of-the-art reinforcement learning (RL) methods are yet to resolve reliably. We address it with three contributions that add just enough structure to guide exploration: Sparse Interaction Edges (SIE) for whole-body manipulation, Interaction Gated Transport (IGT) for sustained transport over long horizons, and a Hinted Object Path (HOP) that resolves collision and search complexity bottlenecks. Across all three, the policy remains free to discover its own manipulation strategy. The method generalises from rigid object to human transport. As a flagship demonstration, we show, to our knowledge, the first bed-to-wheelchair transfer of a medical mannequin by a humanoid robot completed with no human intervention during execution. In the domain of human transfer, we show that a single policy can adapt across substantial variations in body mass, size, and random force perturbations, all while training on a single GPU in under an hour. Our approach offers a pathway to fully RL-driven humanoid manipulation, enabling assistive robots that can help humans. Website: \url{https://elle-miller.github.io/into\_my\_arms/}
    }
  \vskip 2.5em
]

\begin{figure*}[t!]
    \centering
\includegraphics[width=\linewidth]{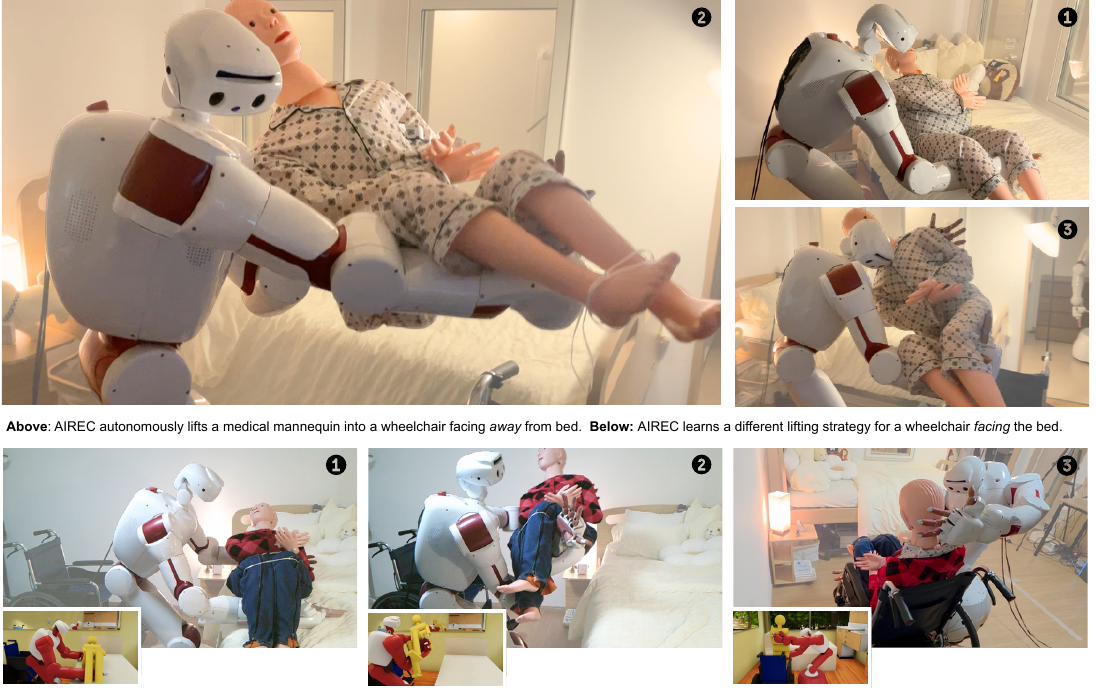}
    \caption{
    \textbf{Autonomous human bed-to-wheelchair transfer on AIREC.} Trajectories from a policy trained entirely in simulation are deployed zero-shot on the physical robot, lifting a medical mannequin (155 cm, 15 kg) from a bed to a wheelchair with no human intervention during execution. Unlike prior humanoid bed-to-chair systems that required a caregiver for key sub-actions \cite{riba}, our full transfer process runs unattended. Two wheelchair placements are shown, with the corresponding simulated rollout inset.
    }
    \label{fig:heroshot}
\end{figure*}

\section*{Introduction}

A robotic agent trained via reinforcement learning (RL) in simulation is free to exploit an embodiment that differs markedly from our own, discover non-obvious strategies, and acquire closed-loop policies that remain robust under substantial scene variation and perturbation. 
Within this paradigm, RL agents have achieved superhuman feats in high-speed drone racing \cite{kaufmann2023champion}, quadrupedal parkour \cite{rudin2025parkour}, and blind in-hand manipulation \cite{miller2025tactilerl}.
Yet, fully RL-driven humanoid manipulation has not seen the same gains. Most humanoid manipulation work instead integrates RL with motion priors or imitation learning, enabling humanoids to lift medium-sized rigid objects \cite{barreiros2025learning, zheng2025embracingbulkyobjectshumanoid, yang2026omniretargetinteractionpreservingdatageneration, rodriguez2026training}. Under pure RL, results remain restricted to bi-manual handovers of small rigid objects \cite{toru} or end-effector-driven box-transfer \cite{zhang2024wococo}. 

We attribute this gap to three intersecting bottlenecks. 
(1) \textit{High state-action dimensionality from coupled robot-object dynamics}: While locomotive agents (drones, quadrupeds) operate with 6–12 degrees of freedom (DoFs) to move themselves, humanoids designed for manipulation often possess 30 or more DoFs and must additionally control the DoFs of an external object. Each DoF exponentially expands the search space and the exploration burden.
(2) \textit{Discontinuous optimisation landscapes from contact-rich interaction}: Standard entropy-based exploration relies on isotropic Gaussian sampling in joint space. In free space, this yields continuous feedback; under tight interaction bounds, the same perturbations instead cause collisions or invalid configurations, making the reward landscape difficult to optimise.
(3) \textit{Long, acyclic task horizons:} Cyclic tasks (e.g., locomotion or in-hand rotation) permit rapid credit assignment over repeated motion primitives. Longer-horizon tasks can instead require non-repetitive sequences of distinct phases (e.g., lifting, transporting, placing), each demanding different skills. 

State-of-the-art RL approaches can sometimes address one or two of these challenges in isolation, but can fail when confronted with this trilemma simultaneously. Closing the gap requires methods that remain general-purpose, rather than being engineered for a single task.
We ground our contribution in a task that embodies all three bottlenecks at once and which carries substantial real-world value: autonomous human bed-to-wheelchair transfer (Figure~\ref{fig:heroshot}). Physical caregiving is an increasingly urgent application for robotics. More urgent, we argue, than the household automation that dominates the field's current attention \cite{aude}. Human mobility is inherently fragile, and demand for care is accelerating. By 2100, nearly a quarter of the global population is projected to be aged 65 or older, exceeding 35\% in nations like Japan \cite{owid-age-structure}. Relying solely on human caregivers to absorb this demand compromises recipient autonomy because independence in daily mobility then depends on another person's availability, while further straining an overwhelmed workforce \cite{msd}. Bed-to-chair transfers are essential for well-being and preventing complications such as pressure injuries \cite{BANKS2024103584}, yet remain among the most physically hazardous tasks for care staff, persisting despite known risks due to staffing shortages and time pressure \cite{noble2018barriers}. Automating this burden does not dehumanise care: it frees caregivers to focus on social and emotional connection, and restores autonomy to recipients \cite{caregiving}. That hope is voiced clearly by Henry Evans, left mute and quadriplegic by a 2002 stroke, who, with Jane Evans, pioneered assistive robotics for disability through Robots for Humanity: \textit{``for the severely disabled, for whom miraculous medical breakthroughs don't seem feasible in our lifetimes, robots are the best hope for significant independence"} \cite{henry}. 

In the assistive domain, RL has so far been limited to single-limb manipulation tasks in simulation, such as lifting an arm or leg \cite{erickson2020assistive, hinckeldey2025assistax, bathing}, falling short of the whole-body, contact-rich, long-horizon transfer we address here.
Beyond its real-world significance, bed-to-chair transfer is a demanding technical testbed. Our humanoid platform AIREC (AI-driven Robot for Embrace and Care) \cite{miyake_deep} has an action space of 35 DoFs and must manipulate an articulated human model with 15 DoFs of its own, squeezing between the human and the bed, lifting and carrying them to the wheelchair, and precisely placing them down within tight insertion tolerances, all without tipping over. With a combined positional state dimension of 50 and control at 30 Hz, the search space spans tens of thousands of dimensions. Robot-human, robot-environment, and human-environment contact discontinuities further complicate the search space.

\begin{figure*}[!t]
       \centering
    \includegraphics[width=0.99\textwidth]{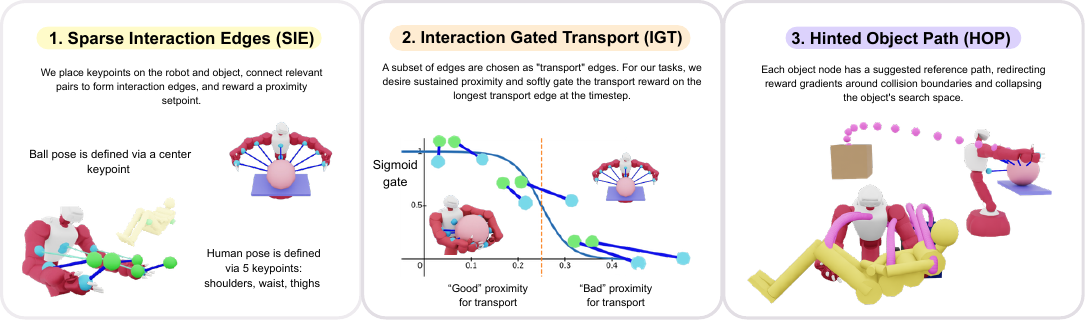}
    \caption{\textbf{Structured reinforcement learning for object transport.} We propose three generalisable methods to guide learning in whole-body, contact-rich, and long-horizon manipulation: Sparse Interaction Edges, Interaction Gated Transport, and Hinted Object Path. These ideas can be integrated into the MDP's observation and reward function without changes to the RL algorithm or training process.}
    \label{fig:method} 
\end{figure*}

We address these bottlenecks through three generalisable ideas to guide exploration (Figure~\ref{fig:method}). First, standard RL formulations feed raw joint configurations \cite{barreiros2025learning}, and sometimes end-effector poses \cite{sferrazza2024humanoidbench, toru}, into the MDP, creating a mismatch with task-space rewards that we hypothesise contributes to poor exploration. We resolve this with Sparse Interaction Edges (SIE), unifying state and reward through keypoints connected across the robot and object, aligning the optimisation gradient with the policy's own observations. Second, standard transport rewards measure only object-to-goal displacement, permitting unconstrained solutions such as throwing an object when sustained proximity is what is actually desired; our Interaction Gated Transport (IGT) softly gates the transport reward on this same SIE structure. Third, jointly discovering robot and object trajectories from scratch creates a prohibitively high-dimensional search, and standard exploration fails wherever agents must manoeuvre around obstacles; our Hinted Object Path (HOP) instead guides exploration with an explicit object path. Across all three, we give RL just enough task structure to guide learning (which spatial relationships matter, how closely the robot must stay engaged, and where the object should travel), while leaving the precise manipulation strategy for the policy to discover, refine, or abandon on its own. 

Together, these contributions solve three whole-body, contact-rich, long-horizon tasks where an ablated RL baseline fails, from transporting a ball into a box to transferring humans. As a flagship demonstration, we show, to our knowledge, the first bed-to-wheelchair transfer of a medical mannequin by a humanoid robot completed with no human intervention (Figure~\ref{fig:heroshot}). Within the human transfer task, we show that a single policy can adapt across substantial variation in human body mass and size.
($115$--$180$\,cm, $11$--$33$\,kg). Training is highly efficient, reaching competitive performance in under an hour  on a single consumer GPU across 2048 parallel environments. By addressing core bottlenecks in high-dimensional robot learning, this work offers a pathway to fully RL-driven humanoid manipulation, enabling assistive robots that can help humans.

\begin{figure*}[!t]
       \centering
    \includegraphics[width=\textwidth]{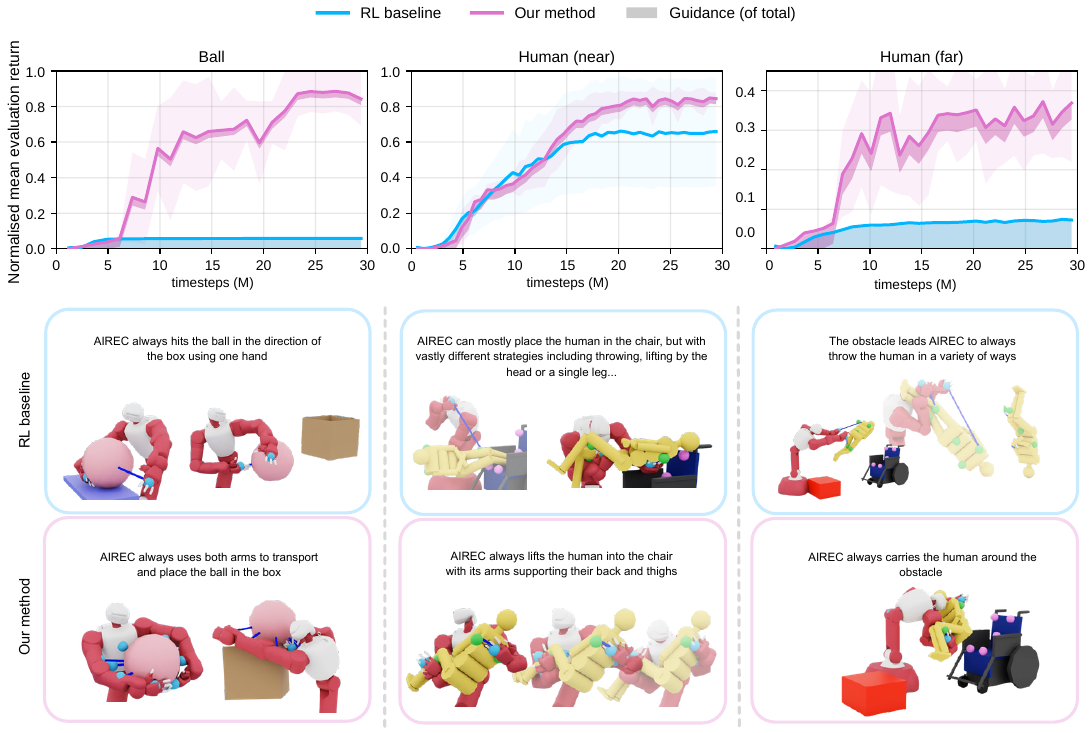}
    \caption{\textbf{Mean evaluation returns and characteristic strategies across 5 seeds.} Top: normalised mean evaluation return with the guidance reward component shaded during training for \emph{Ball}, \emph{Human (near)} and \emph{Human (far)}. Bottom: qualitative strategies of the ablated RL baseline versus our method in each environment.}
    \label{fig:rewards} 
\end{figure*}

\section*{Results}\label{sec:results}

Here we describe our key findings. 
As is common in RL, we report the mean and variance of evaluation returns across different seeds (Figure~\ref{fig:rewards}). 
However, to fully illustrate the effectiveness of our learned policies, we provide per-seed videos on our project website (\url{https://elle-miller.github.io/into\_my\_arms/}).

\paragraph{Setup.} We train our agent using clipped PPO \cite{schulman_proximal_2017} to control AIREC, a 35-DoF mobile humanoid (see Methods). We evaluate three novel environments: \emph{Ball}, a rigid sphere transported from a table into a raised box $2.5$\,m away; \emph{Human (near)}, transferring an articulated human model from a bed into a nearby wheelchair; and \emph{Human (far)}, the same transfer task but over a longer distance with an obstacle in the way. Full agent, environment, and training details are provided in Methods and Appendix.

\paragraph{Ablated RL Baseline.} To our knowledge, no published pure-RL method targets this combination of whole-body control, complex object contact, and long-horizon transport, and prior whole-body approaches typically rely on motion priors or imitation \cite{barreiros2025learning, zheng2025embracingbulkyobjectshumanoid, yang2026omniretargetinteractionpreservingdatageneration, rodriguez2026training}. We therefore compare against an ablated RL baseline that mirrors common design choices from that literature while removing our contributions: the agent observes joint configurations \cite{barreiros2025learning} and end-effector relations \cite{sferrazza2024humanoidbench, toru}, SIE is restricted to the two end-effectors, giving two robot-object edges for the observation and reward, and HOP gated by IGT is replaced with a standard object-to-goal distance reward. Sequential ablations in Figure~\ref{fig:ball} isolate each component's contribution.

\paragraph{Compute.} On a single NVIDIA GeForce RTX 5090, \emph{Ball} trains in about $30$ minutes, while competitive \emph{Human} performance appears within about $60$ minutes. The longer \emph{Human} wall-clock is dominated by articulated-body, contact-rich simulation. Throughput is essentially identical for our method and the RL baseline in each environment, so the extra guidance structure does not measurably increase training cost.

\subsection*{Sparse Interaction Edges enable and guide whole-body manipulation}

Sparse Interaction Edges (SIE) establish spatial correspondence between robot and object keypoints, aiding initial grasping and later guiding transport via the interaction gate. In \emph{Ball}, the single object keypoint sits at the centre, so our SIE formulation leaves the grasp location unguided, giving the policy flexibility in its manipulation strategy. The \emph{Human} environments are designed differently: we deliberately choose object-robot keypoint pairs that result in a more comfortable transfer.

Figure~\ref{fig:rewards}Figure~\ref{fig:ball} shows the effect of enabling whole-body SIE in \emph{Ball}, comparing the first and second panels. The RL baseline agent (first panel) uses its right hand to push the ball behind itself toward the goal, lifting its left arm out of the way; it fails to lift or carry the ball across all five seeds (Figure~\ref{fig:rewards}). We attribute this directly to the MDP formulation: the agent can only ``see'' the end-effector-to-object distance vectors, and is rewarded only for closing those distances, so the optimisation aligns with end-effector motion alone, leaving the rest of the robot's embodiment unused. When we instead let the agent observe additional robot-object keypoint pairs, and reward minimising these edges, it learns to encircle the object with both arms and lift it off the table (second panel). The agent still fails the task, but the shift in strategy is the important result: rather than exploiting a single end-effector, it now manipulates the object using its whole body.

We observe a similar phenomenon in the human environments, with wider variation across seeds: the baseline agent typically throws the human, drops it into its other hand, or lifts by a single thigh or the head, rather than supporting it appropriately (\emph{Human (near)} in Figure~\ref{fig:rewards}; further variation in \emph{Human (far)} is discussed below). These results suggest two primary benefits of SIE. First, it facilitates whole-body manipulation via observation-reward alignment. By perceiving and directly acting on robot-object spatial relationships, the agent no longer has to discover these relationships implicitly in the joint space. Second, it biases the solution toward desirable behaviour through robot-object keypoint design alone, without additional reward engineering.

\subsection*{Interaction Gated Transport enables consistent, but not constraining, long-horizon strategies}

Standard transportation rewards typically measure only object-to-goal displacement, permitting unconstrained solutions such as throwing or pushing. For long-horizon transport tasks, however, sustained proximity to the object is usually what is actually desired. After learning to grasp and lift an object via SIE, we apply Interaction Gated Transport, which acts like an activation function on the transport reward, based on the lengths of a chosen subset of SIE edges.

We see the direct effect of IGT in \emph{Ball} by comparing the second and third panels of Figure~\ref{fig:ball}. The initial grasp and lift are nearly identical, but instead of throwing the ball, the IGT-enabled agent maintains its grasp and carries the ball to the box. The same pattern appears, with greater variation, in the human environments. For the RL baseline in \emph{Human (near)}, the transport strategy differs in every seed, since the unconstrained reward permits any path to the goal. This variation is magnified under the longer horizon of \emph{Human (far)}: seed 1 transports the human upside-down by the ankles; seeds 2 and 4 launch the human into the air, overhead and behind; seed 3 comes closer, but ends up flipping the human $360^\circ$; and only seed 5 succeeds. In contrast, our method transports the human in the same style across every seed in both human environments, with only minor variation, e.g., in \emph{Human (near)} seed 2, the agent's right hand wraps around the left thigh. In contrast, for other seeds it wraps around the upper left calf.
By zeroing the transport reward whenever robot-object keypoint distances exceed a set threshold, IGT prevents early convergence to these degenerate strategies and enables stable, repeatable behaviour over long horizons.

\begin{figure}[t!]
    \centering
\includegraphics[width=\linewidth]{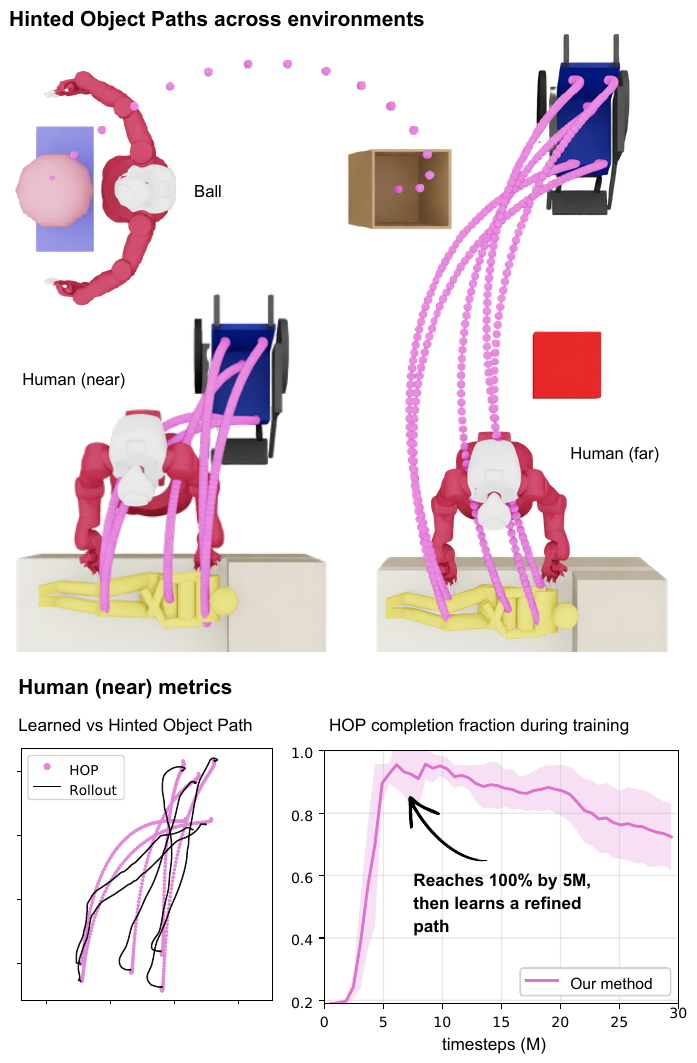}
    \caption{
    \textbf{Hinted Object Paths (HOP) across environments.} Top: HOP waypoints in \emph{Ball}, \emph{Human (near)}, and \emph{Human (far)}. Bottom left: an example learned rollout (black) versus hints (pink) in \emph{Human (near)}. Bottom right: HOP completion rises early in training, then falls as the policy refines a path that still succeeds.
     }
    \label{fig:hop}
\end{figure}

\subsection*{Hinted Object Path resolves collision and search-complexity bottlenecks}

Standard object-to-goal displacement rewards do not scale to contact-rich, high-dimensional transport problems. The gradient of fastest reward accumulation points directly at the goal, which in the presence of obstacles points \emph{into} a collision boundary rather than around it, trapping exploration. The Hinted Object Path instead defines a reward over a full path to the goal, redirecting this gradient around such boundaries. We see this directly in \emph{Ball}, comparing the third and fourth panels of Figure~\ref{fig:ball}: SIE and IGT alone already enable the agent to lift and carry the ball with its whole body, but with an object-to-goal reward the gradient of fastest accumulation points into the box itself, and the agent cannot explore around this collision boundary. With HOP, the object path and thus reward gradient is instead directed up and over the box, leading to task success.

HOP's second benefit, reducing search complexity, grows with object complexity and is clearest in the human environments, where jointly discovering a robot and object trajectory across many object keypoints is itself a high-dimensional search problem, independent of any obstacle. Without HOP, the RL baseline's human keypoints follow wildly different paths across seeds (inserted feet-first, dropped through the upper body, carried upside-down, or spun mid-transport) and in \emph{Human (far)}, the agent resorts to throwing the human in a variety of ways (Figure~\ref{fig:rewards}). With HOP, the agent clears the obstacle. Importantly, the hint is not a rule: keypoints track the reference early, completing the entire path around $5$M timesteps in \emph{Human (near)} before gradually departing from it (Figure~\ref{fig:hop}).
These results show that HOP does not need to be precise to be useful: a coarse object path is enough both to redirect exploration around contact-rich obstacles and to collapse an otherwise vast object search space down to a single reference, after which the agent is free to refine or abandon it. While IGT shapes \emph{how} the robot engages with the object, HOP guides \emph{where} the object travels; the two act on different parts of the problem and compose naturally, together turning an otherwise intractable joint search into one the agent can solve reliably.

\subsection*{Immediate generalisation across diverse human masses and sizes}
\label{sec:results_generalisation}

An implicit benefit of using RL for object transport, rather than task-specific engineered control, is generalisability: a closed-loop policy should adapt to variation in the object itself, not just the robot's own state. Here, we test whether our proposed method retains this benefit after being trained with our additional task structure. We study this using the human transfer task, both because articulated bodies present a harder generalisation challenge than a rigid object, and because human bodies vary substantially in practice: a real-world necessity for any caregiving system. We train different policies under independent variations in human mass ($11$--$33$\,kg) and height ($115$--$180$\,cm) (joint torque limits constrain our physical platform to a maximum payload of $\sim35$\,kg).

\begin{figure*}[!t]
    \centering
    \includegraphics[width=\linewidth]{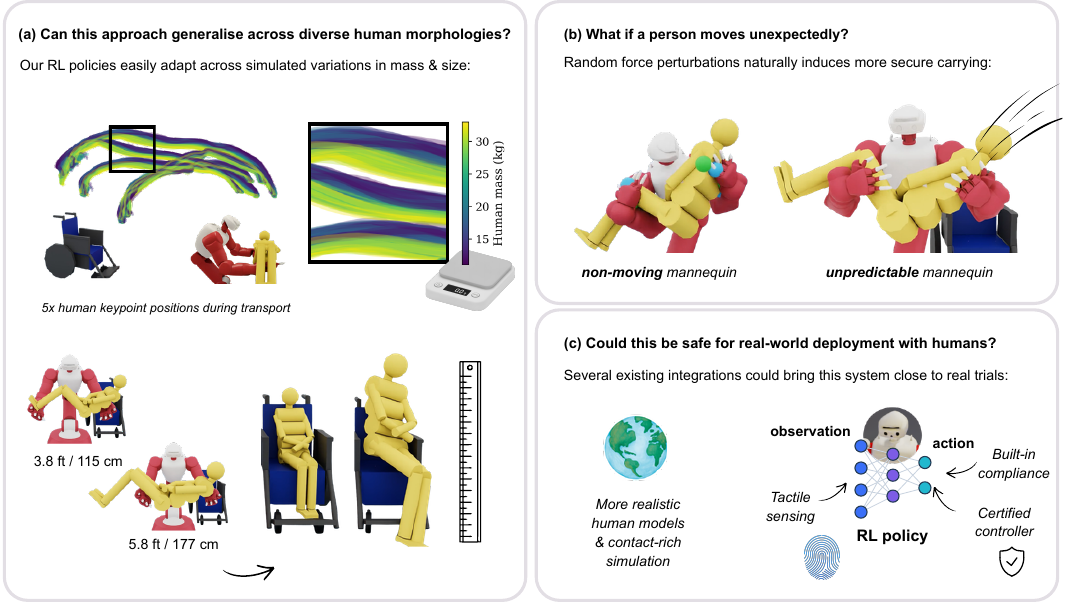}
    \caption{\textbf{Generalisation, robustness, and a path toward safer deployment.}
\textbf{(a)} A single policy adapts across diverse human morphologies in simulation, including mass variation (colour scale) and body size ($115$--$180$\,cm).
\textbf{(b)} Random force perturbations that emulate unexpected recipient motion induce a tighter, more secure carry compared with an unperturbed mannequin.
\textbf{(c)} Several existing ingredients could help bring such a systems closer to real human trials.}
    \label{fig:generalisation}
\end{figure*}

In Figure~\ref{fig:generalisation}(a), across 1,000 rollouts spanning this mass range,  we can see that the policy largely guides the human along a common path regardless of mass, with some variation: lighter bodies follow a slightly elevated trajectory and correspondingly sharper robot joint inflexions, indicating tighter trajectory tracking.
A trained policy is also able to adapt across human height, since representing the body and task goals via five keypoints endows the observation space with structural geometric invariance. The main challenges instead came from physical scene alignment: the workspace and initial robot configuration were tailored to the $155$\,cm mannequin, leaving suboptimal starting placements at extreme heights (e.g., the $115$\,cm mannequin's feet resting on the robot's hand), and the compact Japanese wheelchair frame struggled to accommodate individuals near $180$\,cm. Despite these alignment and workspace constraints, the agent recovers and completes the transfer task in most cases, demonstrating generalisation across substantial variation in human morphology without any additional training.

\subsection*{Robust closed-loop behaviour to unpredictable object motion}
\label{sec:results_reactive}

Real-world transport rarely goes exactly to plan: objects can shift, slip, or be jostled mid-transport, and a robust policy must react rather than blindly follow a fixed trajectory. This is especially true for human transfer, where care-recipients with motor impairments may experience involuntary spasms or tremors. To evaluate the agent's capacity to react to such disturbances, we apply random external forces to the target object in both the \emph{Ball} and \emph{Human} environments: at each timestep, a force sampled uniformly from $[300, 500]$\,N is applied with probability $p=0.1$ to the centre of the ball or the chest of the human. 

\begin{figure*}[!t]
       \centering
    \includegraphics[width=0.8\textwidth]{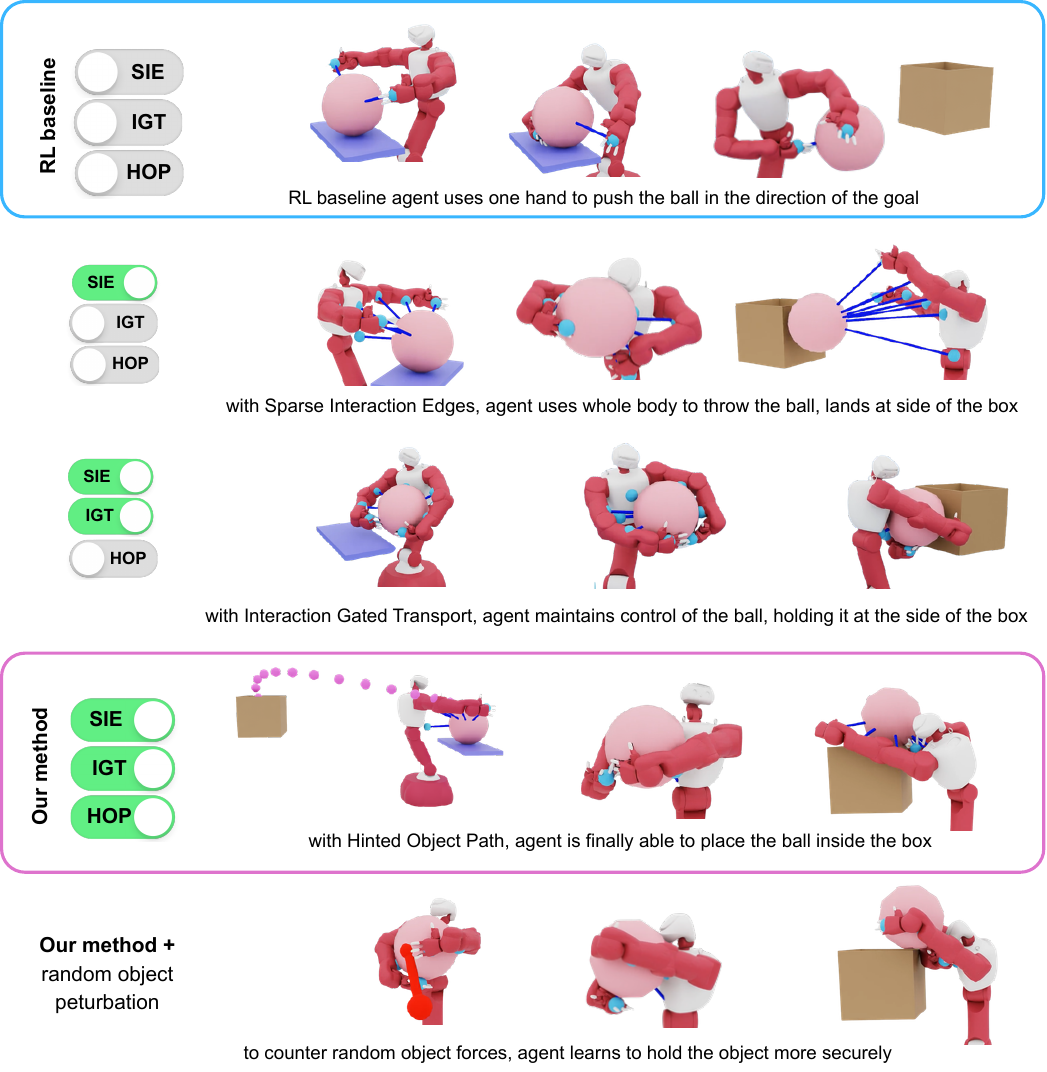}
    \caption{\textbf{Method ablation on the \emph{Ball} environment.} Starting from the ablated RL baseline, SIE, IGT, and HOP are enabled in sequence to highlight the direct effect of each method on the behaviour rollout. In the final panel, the agent is trained with random force perturbations applied to the object (red arrows). Note that the actual HOP waypoint tolerance is set to $0.5$\,m, but the waypoints are sized down for visualisation.}
    \label{fig:ball} 
\end{figure*}

Introducing these perturbations prompted the agent to discover a distinct, more robust manipulation strategy while preserving task success. In \emph{Ball} (final panel, Figure~\ref{fig:ball}), the unperturbed agent executes a symmetrical ``hug'' grasp, placing both hands together and supporting the ball atop its forearms; under perturbation, it instead learns an asymmetrical grasp (right hand underneath for primary support, left hand around the centre for stabilisation) maintained throughout transport. We observe an analogous adaptation in \emph{Human}. Whereas unperturbed rollouts rest the left hand at the centre of the back with the right arm at a right angle (Figure~\ref{fig:heroshot}), the perturbed policy wraps the left hand over the human's left shoulder, drawing the right hand much closer to the chest and securing the human closer to its torso (Figure~\ref{fig:generalisation}(b)).

Across both domains, random object perturbations during training prompt the emergent discovery of tighter, more secure grasps without any explicit reward, demonstrating RL's capacity for robust, closed-loop manipulation under unpredictable object dynamics.

\subsection*{First bed-to-chair transfer of a medical mannequin without intervention}
\label{sec:results_sim2real}

As a flagship demonstration, we replayed trajectories from our trained policy, open-loop, on the physical AIREC robot in a replica bedroom with an adjustable-height hospital bed and standard wheelchair (Figure~\ref{fig:heroshot}), using a medical training mannequin ($155$\,cm, $15$\,kg) with realistic joint motion. Despite several sim-to-real challenges (Appendix~\ref{app:sim2real}), AIREC successfully lifts the mannequin from the bed, carries it to the wheelchair, places it down, and disengages. We test two wheelchair placements, facing the head of the bed and facing away at the foot, both requiring a $90^\circ$ rotation in addition to translation (Figures~\ref{fig:heroshot}). To our knowledge, this is the first bed-to-wheelchair transfer of a medical mannequin by a humanoid completed without a caregiver in the loop.

Closing the sim-to-real gap required addressing four challenges, detailed in Appendix~\ref{app:sim2real}: an unmodelled real-world impedance controller diverging from our idealised simulated PD law, matching the real robot's strict stability monitoring, which halts execution if the Zero-Moment Point exits its support polygon; eliminating large contact forces on the wheelchair that would tip it in reality but were invisible to a fixed simulated chair; and precisely calibrating the simulated scene, including the mannequin's shape, to match the physical setup.
Even so, replaying the same trajectory did not always yield the same outcome, and we observed failures alongside successful rollouts. This demonstration shows that the learned strategy is physically realisable; it is not a full closed-loop hardware validation study.

\section*{Discussion}
SIE, IGT, and HOP supply just enough structure for whole-body, contact-rich, long-horizon transport where an ablated RL baseline fails. Applying our method to a new task requires a small set of designer choices: robot, object, and goal keypoints; interaction edges that gate transport; a coarse, hinted object path; and completion tolerances. We see this as a light design burden compared with gathering teleoperated or human demonstrations for every new object, scene, and goal.
Critically, the guidance is a hint instead of a rule: the SIE/IGT/HOP reward terms contribute only a small fraction of total return (e.g., 5-10\%, Figure~\ref{fig:rewards}), so the policy remains free to adopt an alternative strategy for the guidance terms are zero. In practice, it often approximately follows the suggested engagement and path early on, then departs when beneficial.

One may ask why a general-purpose humanoid controlled by a learned RL policy is needed, rather than alternatives such as a robotic hoist or bed-mounted transfer system. We see a general-purpose humanoid as one attractive path alongside specialised devices such as robotic wheelchairs and exoskeletons. One mobile platform can share sensing and control advances across many caregiving tasks, so hardware and software effort need not be rebuilt for each new assistive function. In addition, a free-moving humanoid can, in principle, accompany a person beyond the bedside into settings that remain largely inaccessible today, such as rugged terrain or commercial air travel \cite{inaccessible}. Thus, we build on a long line of general-purpose humanoid caregiving robots. Between 2006 and 2015, a series of Japanese humanoids demonstrated the first bed-to-chair transfers of humans (RI-MAN \cite{riman}, RIBA \cite{riba}, TWENDY-ONE \cite{twendyone}). Those systems were semi-autonomous: execution relied on a pre-defined trajectory, with a caregiver in the loop to perform key sub-actions (e.g., lifting the legs) and make fine adjustments for individual human properties (e.g., height).

Achieving full autonomy raises a related question of learning paradigm. More recently, model-based approaches have generated desired contact forces for autonomous assistance via whole-body quadratic programming, for sit-to-stand \cite{sit2stand} and arm repositioning \cite{anastasia}, and via motion planning for supine-to-lateral repositioning \cite{misa}. These approaches offer valuable safety guarantees, but their reliance on an accurate model makes them harder to apply when the human, an unpredictable agent rather than a passive object, is difficult to plan for \cite{billard_trends_2019}. Imitation learning (IL) trades this modelling burden for a different one: recent approaches replicate teleoperated repositioning manoeuvres, such as sitting up in bed \cite{miyake2025dualarmmotiongenerationrepositioning}, offering a simpler route to competent behaviour, though by construction reproducing only a fixed demonstrated trajectory limits their ability to react to unpredictable human behaviour at test time. RL instead offers closed-loop, dynamically adaptive policies well suited to this unpredictability, while also exploiting a robotic caregiver's high-torque actuators, non-biological ranges of motion, and superhuman stability, rather than mimicking human caregiver technique constrained by spinal vulnerability \cite{BANKS2024103584}. We see these paradigms as complementary rather than competing: combining RL's adaptability with the safety guarantees of model-based control, or the sample efficiency of imitation learning, is a promising future direction.

Having motivated both choices, a further question is how close this brings us to deploying with an actual human rather than a mannequin (Figure~\ref{fig:generalisation}(c)). Two barriers remain: motor strength and safety, and resolving them defines our main directions for future work. Holding a $50$\,kg person (and own $25$\,kg arm) at $0.5$\,m requires each shoulder motor to supply approximately $375$\,Nm, above AIREC's current rated $220$\,Nm; stronger motors exist, so the limit is engineering rather than physics. Safety is more fundamental as a typical neural-network policy offers no certificate that small shifts in state will not produce unsafe  actions. Progress on certifiable robustness for neural control in safety-critical settings \cite{nn_safety} is one path toward closing that gap. 
More fundamentally, we view whole-body tactile sensing as imperative for future iterations, both to relax the framework's dependence on precise object keypoint perception and to ensure human safety, comfort, and robustness to dynamic motion, slippage, and occlusion during real physical contact. Compliant, force-limited whole-body control \cite{lu2025gentlehumanoidlearningupperbodycompliance, margolis2025softmimic} is a complementary path toward safer interaction even before such sensing is mature. Resolving the RL trilemma on this caregiving testbed is a step toward assistive autonomy; human deployment still requires the barriers above to be addressed.

\section*{Methods}

We introduce three generalisable ideas to guide learning toward effective solutions in whole-body, contact-rich, long-horizon RL: Sparse Interaction Edges (SIE), Interaction Gated Transport (IGT), and Hinted Object Path (HOP), plus Analytical Bandwidth Computation (ABC) for reward scaling. All four act only through the MDP's observation and reward and are compatible with any RL algorithm. The total reward is the sum of our guidance reward, and a more typical task reward:

\begin{equation}
    R_\text{total} = R_\text{guidance} + R_\text{task}
    \label{eq:reward}
\end{equation}
\begin{equation}
    R_\text{guidance} = \underbrace{R_{\text{SIE}}}_{\text{reach}} 
    + \underbrace{\sigma_{\text{IGT}} R_\text{HOP}}_{\text{transport}}
    \label{eq:reward}
\end{equation}
$R_{\text{SIE}}$ drives reaching, and $\sigma_{\text{IGT}} R_\text{HOP}$ drives gated transport along a hinted path. The task reward covers fine object placement, robot reset, and sparse success or stability terms (Appendix~\ref{app:task_reward}).

\subsection*{Sparse Interaction Edges (SIE)}
\label{sec:method_sim}

Standard RL baselines typically observe joint configurations \cite{barreiros2025learning} and sometimes end-effector poses \cite{sferrazza2024humanoidbench, toru}, paired with task-space rewards such as object-to-goal or end-effector-to-object distance. We hypothesise that this representation is insufficient for whole-body manipulation, leading to poor exploration.
Several recent works build explicit spatial representations between robot and object, but for purposes distinct from resolving this exploration problem directly. OmniRetarget \cite{yang2026omniretargetinteractionpreservingdatageneration} constructs an interaction mesh over robot, object, and terrain keypoints, dating back over a decade to early interaction-mesh-based motion adaptation \cite{nakaoka}, but uses it only to improve motion-retargeting quality. The RL policy itself remains blind to this structure, observing basic proprioception and tracking a reference trajectory precisely. AnyBody \cite{li2026anybodyfreeformwholebodyhumanoid} similarly guides whole-body control via a sparse set of robot keypoints, but for motion tracking alone, with no object or manipulation task. Yuan et al. \cite{yuan_rescue} instead represent robot-human interaction through a topology-based Writhe matrix, demonstrating sim-to-real transfer of a dual-arm holding policy for swimmer rescue. In contrast, Zheng et al. \cite{zheng2025embracingbulkyobjectshumanoid} perceive robot-object interaction through a latent neural signed-distance representation computed by a pre-trained network, relying on motion priors to accelerate learning in long-horizon, contact-rich tasks. 

None of these representations is used to directly shape both what the policy observes \emph{and} what it is rewarded for.
We instead resolve this representation mismatch by unifying state and reward through Sparse Interaction Edges: a lightweight, hand-specified set of keypoints on the robot and manipulated object, connected by 3D spatial edge vectors, requiring no reference motion, motion priors, or latent perception. Formulating the reward as a scalar field over these edge lengths aligns the optimisation gradient with the policy's observation features, accelerating convergence in whole-body tasks. The resulting low-dimensional topology also acts as a compact map, projecting the high-dimensional configuration space onto a task-relevant manifold.

\paragraph{Construction.} Let $N_r$, $N_o$, and $N_g$ be the numbers of robot, object, and goal keypoints, with $N_g = N_o$. Each robot keypoint maps to one object keypoint, and several robot keypoints may share an object keypoint, e.g., both hands mapping to a single box centroid. Keypoint placement is left to the designer: keypoints can sit anywhere on or within the robot or object, and object keypoints need not be uniformly distributed; the topology can instead be tailored to task-relevant geometry such as contact points, grasp regions, or articulated structure. This construction scales from a minimal single-keypoint-per-set topology to a dense interaction mesh, trading representational richness against observation dimensionality. These robot-object edge vectors form the basis of both the agent's observation space and its reward function.
The robot's keypoints are first drawn to the corresponding object keypoints via the reach reward $R_{\text{SIE}}$: the mean normalised distance across all robot-object keypoint pairs. Euclidean distances are mapped to $(0,1]$ via a hyperbolic tangent transform (Eq.~\ref{eq:tanh}), giving a smooth, non-vanishing gradient with a ``funnel" effect that steepens near the target. An offset $b$ ensures maximum reward is attained at the object's boundary (e.g., its radius) rather than its centre; the scale $k_\text{SIE}$ is derived by ABC.

\begin{equation}
\mathcal{D}(d, k) = 1 - \tanh \left( \frac{d}{k} \right)
\label{eq:tanh}
\end{equation}
\begin{equation}
    R_{\text{SIE}} = \frac{1}{N_r}\sum_{i=1}^{N_r} \mathcal{D}\left(d_i^{\text{SIE}} - b,\ k_{\text{SIE}}\right)
    \label{eq:r2o}
\end{equation}

\paragraph{Implementation.} We use eight robot keypoints (mobile base, chest, upper arms, forearms, and hands). For \textit{Ball}, one object keypoint at the centre receives all robot mappings. For \textit{Human}, we find that five keypoints are sufficient to capture pose (shoulders, thighs, and waist). We map robot hands to the human's left shoulder and thigh, and robot upper/lower arms to the human's right shoulder and thigh (Figure~\ref{fig:method}). 

\subsection*{Interaction Gated Transport (IGT)}
\label{sec:method_igt}

The primary objective of pick-and-place is to translate each object keypoint to a unique goal keypoint, and standard transportation rewards accordingly measure only object-to-goal displacement (e.g., \cite{barreiros2025learning, sferrazza2024humanoidbench, toru}). We found that ungated object-to-goal rewards produced degenerate policies that used only part of the robot, threw the object, or pushed it. Instead, we wanted the robot to stay close to the object throughout transport. 
We therefore introduce Interaction Gated Transport (IGT), which applies a soft gate to the transport reward using the same SIE-based formulation, contingent on proximity rather than contact. This keeps the observation and reward spaces aligned, requires no reference motion, and makes transport contingent on sustained whole-body interaction rather than on object displacement alone.

A related line of work constrains whole-body manipulation via explicit contact goals, rather than leaving the manner of contact unconstrained as above. WoCoCo \cite{zhang2024wococo} decomposes loco-manipulation into sequential contact stages with pre-defined contact goals (e.g., the side of a box); Lin et al. \cite{toru} similarly specify desired contact locations for dexterous fingertips, with an associated contact reward; and MaskedManipulator \cite{tessler2025maskedmanipulator} identifies which body parts should be in contact and rewards proximity to a reference motion's contact configuration. Each of these requires a well-defined notion of contact, e.g., a location on the object surface, which is unavailable in tasks like \emph{Ball}, where the sole object keypoint sits at an uncontactable centre.

\paragraph{Construction.} We gate the object transport reward with a sigmoid, which acts like an activation function; we choose a sigmoid over a binary gate to give smoother gradients. Let $d = \max_{j \in J} d_j^{\text{SIE}}$ denote the maximum SIE edge length within a chosen subset $J$ of robot keypoints. The gate is then:

\begin{equation}
    \sigma_\text{IGT} = \frac{1}{1 + e^{(d - d_0)/k}}
\end{equation}
which rises toward 1 as those edges shorten and unlocks $R_\text{HOP}$.

\paragraph{Implementation.} For \emph{Ball}, $J$ is all edges and $d_0 = 0.5$ m. For Human, $J$ includes the hands and lower arms and $d_0 = 0.3$ m, requiring the robot to keep the human in its arms during transport.

\subsection*{Hinted Object Path (HOP)}
\label{sec:method_bezier}

Jointly discovering robot and object trajectories is high-dimensional, and endpoint-only rewards often drive the agent into obstacles. We hypothesise this is a local-minima problem: an endpoint-based reward directs the agent along the direction of steepest reward accumulation, which in free space coincides with the direct path to the goal, but in contact-rich settings with obstacles, or narrow openings, this direction frequently points \emph{into} a collision boundary rather than around it.

Nearly every RL method for whole-body manipulation that uses a reference, whether joint states, robot keypoints, or end-effector waypoints, guides the robot's own motion with it \cite{barreiros2025learning, rodriguez2026training, iwata2026robustbrachiationlifesizeddualarm, tessler2025maskedmanipulator, yang2026omniretargetinteractionpreservingdatageneration, li2026anybodyfreeformwholebodyhumanoid}. To the best of our knowledge, no prior work in the domain of RL-driven humanoid manipulation provides a standalone \emph{object's} trajectory as a learning reference.
We propose the Hinted Object Path, an explicit hint trajectory for the object itself from start to goal, for each object-goal pair (Figure~\ref{fig:hop}). 
The reward is defined over this full path to the goal rather than the object's final pose alone, redirecting the gradient of fastest accumulation around collision boundaries rather than into them. Any trajectory generation method is compatible with HOP.

\paragraph{Construction.} Each object keypoint has a path of $N_\text{HOP}$ waypoints with tolerance $\delta_\text{HOP}$. The observation includes distances to the next waypoints and path completion. The reward combines path tracking and progress. The waypoints have to be acquired in order.

\begin{equation}
    R_\text{tracking} = \frac{1}{N_o}\sum_{m=1}^{N_o} \mathcal{D}\left(d_m^{\text{HOP}},\ k_\text{tracking}\right)
\end{equation}
\begin{equation}
    R_\text{progress} = \frac{1}{N_o}\sum_{m=1}^{N_o} \mathbbm{1}\!\left[d_m^{\text{HOP}} < \delta_\text{HOP}   \right]
\end{equation}
\begin{equation}
    R_\text{HOP} = R_\text{tracking} + \lambda R_\text{progress}
    \label{eq:hop}
\end{equation}
with $k_\text{tracking}$ from ABC and $\lambda$ chosen so cumulative progress is roughly three-fold the tracking. Final placement precision is defined in the task reward $R_\text{task}$ (Appendix~\ref{app:task_reward}
).

\paragraph{Implementation.} We instantiate paths as quartic Bézier curves. The start and end points of the curves are the initial and target human poses respectively. The control points were chosen by visualising and refining the paths in a 3D view of the scene (further details in Appendix~\ref{app:bezier}).  For Ball, $N_\text{HOP}=10$ and $
\delta_\text{HOP} = 0.5$ m; for Human, $N_\text{HOP}=100$ and $\delta_\text{HOP} = 0.25$ m.

\subsection*{Analytical Bandwidth Computation (ABC)}
\label{sec:method_abc}

For all dense reward components computed via the hyperbolic tangent transform $\mathcal{D}$, we normalise distances and control gradient steepness via a bandwidth parameter $k_\mathcal{D}$. The disadvantage of this formulation is that $k_\mathcal{D}$ requires careful tuning to ensure the reward is meaningfully above zero at the initial distance -- too large a $k_\mathcal{D}$ yields a vanishing gradient at onset, while too small saturates the reward prematurely. This is typically tuned empirically, which is time-consuming. Recent work proposes bi-level optimisation to find the optimal bandwidth \cite{xie2025kungfubot}. Instead, we propose Analytical Bandwidth Computation (ABC): rather than hand-tuning $k_\mathcal{D}$, we specify a desired initial reward value $r_0$ at a known initial (offset) distance $\delta$, such that $\mathcal{D}(\delta, k_\mathcal{D}) = r_0$, and solve for $k_\mathcal{D}$ directly:
\begin{equation}
    k_\mathcal{D}(\delta, r_0) = \frac{2\delta}{\ln\left( \frac{2 - r_0}{r_0} \right)}
    \label{eq:abc}
\end{equation}

We use $r_0 = 0.05$ everywhere. For SIE, $\delta$ is the initial mean robot–object distance; for tracking and precision, $\delta$ is $\delta_\text{HOP}$; for reset, $\delta$ is estimated from post-placement joint error averaged over 100 samples.

\subsection*{Experimental setup}
\label{sec:environments}

\paragraph{Environments.} We evaluate our agent in three environments, \emph{Ball}, \emph{Human (near)}, and \emph{Human (far)}, built in Isaac Lab \cite{nvidia2025isaaclabgpuacceleratedsimulation}, with the control policy running at $30$\,Hz and the physics engine at $120$\,Hz. In \emph{Ball}, the object is a rigid sphere, $50$\,cm in diameter, with randomised mass between $2.5$--$7.5$\,kg, transported from a tabletop surface into a raised box more than $2.5$\,m away. In \emph{Human (near)} and \emph{Human (far)}, the object is an articulated human model composed of spheres and cylinders, with 15 revolute joints (head, neck, chest, shoulder pitch and roll, elbows, hip pitch and roll, knees), a height of $155$\,cm, and randomised mass between $11$--$33$\,kg, transferred from a bed into a wheelchair; \emph{Human (far)} additionally places the wheelchair at a greater distance, with an obstacle in the way.

\paragraph{Agent.} Observations combine proprioception (joints, torques, command errors, centre of mass, planar ZMP) with SIE, IGT, and HOP features (Table~\ref{tab:observations} in Appendix~\ref{app:obs_control}). Actions are 35-DoF joint-position deltas for AIREC (head, torso, arms, hands, omnidirectional base), slew-limited and smoothed, then tracked by a low-level PD controller (gains and limits in Appendix). Episodes use environment-specific warm-start poses with randomised base, object, mass, and goal; they terminate on tip-over, dropped object, excessive displacement, or numerical failure.

\subsection*{Resources} This research was done on a single NVIDIA RTX5090 GPU. This work was supported by JST Moonshot R and D, Grant No. JPMJMS263C.

\subsection*{Supplementary information}
All resources (code, logs, plot generation) can be found on the project website.

\subsection*{Author contributions}
E.M. led the conceptualisation, algorithm design, and methodology, developed the simulation environment, trained the policies, and conducted the experimental evaluation and real-world trials, and drafted the manuscript. T.T. proposed the use of soft gates in place of hard staged rewards. B.S. proposed bypassing explicit object path discovery, which led to the formulation of HOP. R.Al. contributed to the implementation of HOP and real-world success for the bed-to-chair task and, together with Y.W., investigated the robot controller to support sim-to-real transfer. T.M. led the experimental support for physical robot experiments, with additional support from R.Al., T.T., and Y.W. R.An., O.M.A. and S.V. contributed to conceptualisation and supervised the project. S.V. additionally administered the project. Resources were provided by T.M., S.S. and T.O. Funding was acquired by N.S., S.S., T.O. and S.V. All authors contributed to reviewing and editing the manuscript. 

\bibliographystyle{unsrt} 
\bibliography{references}

\appendix

\section{Sim-to-real transfer}
\label{app:sim2real}

To successfully transfer rollouts from trained policies to the physical AIREC robot, we had to overcome several sim-to-real challenges.

\begin{figure}[t!]
    \centering
\includegraphics[width=\linewidth]{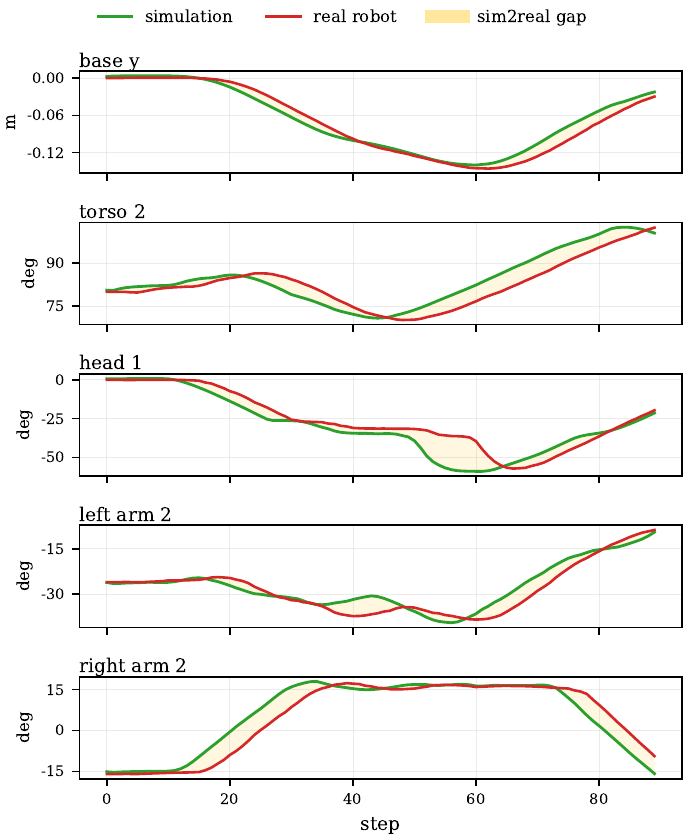}
    \caption{Joint positions and joint torques for a single bed-to-chair rollout in Isaac Lab and on the real robot. Torques are shown for arm joints only (the joints with torque sensors). Joint positions track the commands with a small lag. Torque profiles differ substantially because simulation uses an ideal PD law, whereas the robot uses cascading impedance controllers (exact equation unknown).}
    \label{fig:joint_comparison}
\end{figure}

\textit{1) Controller mismatch.}
Simulation relied on an ideal PD torque law ($\tau = K_p(\theta^\text{cmd}-\theta)-K_d\dot{\theta}$), whereas the physical robot operates under an unmodelled impedance controller. This mismatch produced divergent torque and joint position signals that escalated with joint velocity (Figure~\ref{fig:joint_comparison}). We mitigated the mismatch by replaying the trajectories at a reduced frequency of $10$\,Hz ($1/3\times$ the training frequency).

\textit{2) Stability monitoring.}
The real robot has strict stability safety monitoring to prevent tipping over. An emergency stop triggers if the Zero-Moment Point (ZMP) exits a $12 \times 12$\,cm$^2$ support polygon. We match this behaviour in simulation via a termination condition and a $-1000$ penalty, and add the planar ZMP, centre-of-mass position, and acceleration to the observation so the agent can implicitly learn this constraint.

\textit{3) Wheelchair collision avoidance.}
Fixing the wheelchair position in simulation allowed policies that exerted large forces on the chair to aid placement, which would tip the physical chair in the real world. We eliminated this behaviour by inflating the wheelchair's collision meshes, which removed robot-chair contact in the real world.

\textit{4) Scene alignment.}
We calibrated the simulated environment (bed, wheelchair) to mirror the physical objects. While we trained with position randomisations, the object positions for open-loop replay had to match the physical scene. For the mannequin, we measured each link and approximated its shape with cylinders.

\textit{5) Safety stops from torque and joint-position limits.}
The real robot enforces hard safety stops when commanded torques or joint position errors exceed rated limits. These stops occurred frequently during human placement, where the combination of the human's weight and the precision required to lower them into the wheelchair pushed several joints close to their limits. We could not overcome this limitation directly.

\textbf{Physical setup.}
We conducted trials in a replica bedroom, with an adjustable-height hospital bed and a standard wheelchair (Figure~\ref{fig:heroshot}). We used a medical training mannequin made from PVC with realistic joint motions, height $155$\,cm and weight $15$\,kg, dressed in jeans and a jumper. To reduce complexity and sim-to-real mismatch, we loosely wrapped the wrists and knees together. At present, workspace ergonomics constrains deployment. Reaching down to a standard $45$\,cm-high wheelchair seat forced a trade-off between grasp security and balance, which taller, adjustable seating (already available in consumer power wheelchairs, $50$--$80$\,cm) would relax.

\section{Hinted Object Path construction}
\label{app:bezier}

Methods instantiate HOP with quartic B\'ezier curves. A quartic B\'ezier curve is defined by five control points in $\mathbb{R}^3$, $P_0, P_1, P_2, P_3, P_4$, and parameterised by $t \in [0,1]$:
\begin{multline}
    B(t) = (1-t)^4 P_0 + 4(1-t)^3 t\, P_1 + 6(1-t)^2 t^2\, P_2 \\
    + 4(1-t)\, t^3 P_3 + t^4 P_4.
    \label{eq:bezier}
\end{multline}

The curve starts at $P_0$ and ends at $P_4$, set to the object's initial and goal positions, respectively. The intermediate control points shape the trajectory: the curve leaves $P_0$ heading toward $P_1$, and arrives at $P_4$ from the direction of $P_3$. To avoid control points scaling with the number of object keypoints, we place control points on the central object keypoint only, and derive the remaining keypoints' control points from the object's configuration at the relevant path endpoint. Offsets for $P_1$ and $P_2$ are taken relative to the central keypoint in the object's initial configuration (e.g., a human lying on the bed). The offset for $P_3$ is taken relative to the central keypoint in the object's goal configuration (e.g., a human seated in the chair). 

\section{Task reward}
\label{app:task_reward}

The task reward in Eq.~\ref{eq:reward} is
\begin{equation}
    R_\text{task} = R_\text{precision} + R_{\text{reset}} + c_\text{bonus} - c_\text{stability}.
\end{equation}
$R_\text{precision}$ refines object placement once HOP has brought the object within its coarse tolerance $\delta_\text{HOP}$. $R_\text{reset}$ returns the robot to a default joint configuration $\theta^\text{reset}$ once the object is successfully placed. The reset reward is computed over a joint subset $\mathcal{S}\subset\{1,\dots,N_\theta\}$ corresponding to the torso, arms, and head. Restricting $\mathcal{S}$ this way lets the base move freely to disengage, and avoids the high-dimensional hand DoFs dominating the error. We additionally grant two sparse bonuses $c_\text{bonus}$, once all object keypoints reach their goals and again once the robot returns to its default posture with the object still placed, and a sparse stability penalty $c_\text{stability}$ to discourage tipping.

Rewarding placement precision at every timestep proved exploitable: the agent raced to the goal to accumulate reward for longer at the expense of precision. We instead reward progress in a dense signal $u_t = \mathcal{D}(\cdot, k)$, using its change $r_t = u_t - u_{t-1}$ rather than its raw value. Summed over an episode, this telescopes to the net improvement, independent of the path taken:
\begin{equation}
    R = \sum_{t=1}^{T} r_t = u_T - u_0.
    \label{eq:telescope}
\end{equation}
For precision, let $u_t^\text{precision} = \frac{1}{N_o}\sum_{j=1}^{N_o}\mathcal{D}(d_j^\text{o2g}(t),\ k_\text{o2g})$. Then
\begin{equation}
    R_\text{precision} = u_T^\text{precision} - u_0^\text{precision}.
    \label{eq:precision}
\end{equation}

For reset, let $u_t^\text{reset} = \mathcal{D}(\text{MSE}_\theta^\mathcal{S}(t),\ k_\text{reset})$, with $k_\text{reset}$ from ABC, and
\begin{equation}
    \text{MSE}_\theta^\mathcal{S} = \frac{1}{|\mathcal{S}|}\sum_{i\in\mathcal{S}}(\theta_i - \theta_i^{\text{reset}})^2.
\end{equation}
The reset signal must be suppressed whenever the object is not at the goal; otherwise, the agent can accrue reset reward by pushing the object aside during retreat. We introduce a precision gate $\sigma_\text{precision}$, identical in form to IGT's sigmoid:
\begin{equation}
    \sigma_{\text{precision},t} = \frac{1}{1 + e^{(d^\text{o2g}_t - d_0^\text{o2g})/k_\text{o2g}}}, \qquad d^\text{o2g}_t = \max_j d_j^\text{o2g}(t).
\end{equation}
The reset reward is then
\begin{equation}
    R_\text{reset} = \sum_{t=1}^{T} \sigma_{\text{precision},t}\, u_t^\text{reset}.
    \label{eq:reset}
\end{equation}

\section{Observations and actions}
\label{app:obs_control}

\begin{table*}[ht]
\centering
\caption{Agent observation space.}
\label{tab:observations}
\small
\begin{tabularx}{\textwidth}{@{} l l l X @{}}
\toprule
\textbf{Category} & \textbf{Component} & \textbf{Symbol} & \textbf{Description} \\ \midrule
\textbf{Proprioception} & Joint states  & $\hat{\theta}, \hat{\dot{\theta}}, \hat{\tau}, \Delta\mathbf{a}$ & Norm. joint pos., vel., torques, and action deltas. \\
                  & Control errors  & $\mathbf{e}^{\theta}, \mathbf{e}^{\tau}$ & Position $(\theta_{\text{cmd}} - \theta)$ and torque clipping errors. \\
                  & Dynamics        & $\mathbf{x}_{\text{com}}, \ddot{\mathbf{x}}_{\text{com}}, p_x, p_y$ & Centre-of-mass position, acceleration, and planar ZMP. \\ \addlinespace[0.5em] \midrule
\textbf{SIE} & Robot keypoints     & $\{\mathbf{x}_i^r, \dot{\mathbf{x}}_i^r, \mathbf{g}_i\}_{i=1}^{N_r}$ & Cartesian pos., vel., and 6D rotation. \\
                  & Object keypoints    & $\{\mathbf{x}_j^o, \dot{\mathbf{x}}_j^o\}_{j=1}^{N_o}$ & Cartesian pos. and vel., relative to robot base. \\
                  & Goal keypoints      & $\{\mathbf{x}_j^g\}_{j=1}^{N_o}$ & Cartesian pos. of target, $N_g = N_o$. \\
                  & SIE edges & $\mathbf{d}^\text{SIE}, \dot{\mathbf{d}}^\text{SIE}, d^\text{SIE}, \mathcal{D}(d^\text{SIE})$ & Robot-object edge vectors, speeds, scalars, and normalised scalars. \\ \addlinespace[0.5em]
                  \midrule
\textbf{IGT} & Interaction gate & $\sigma_{\text{IGT}}$ & Soft-gate activation for object transport. \\
\midrule
\textbf{HOP}
                  & Waypoint distances   & $\mathbf{d}^{\text{HOP}}_{m,1:h}, d_m^{\text{HOP}}$ & Distance to next $h$ waypoints and scalar distance to the active waypoint, per object keypoint $m$. \\
                  & Path completion & $\phi_m$ & Path completion fraction, per object keypoint $m$. \\
                  \midrule
\textbf{Precision/Reset} & Object-goal edges &  $\mathbf{d}^\text{o2g}, \dot{\mathbf{d}}^\text{o2g}, d^\text{o2g}, \mathcal{D}(d^\text{o2g})$ & Object-goal edges as for SIE. \\
& Reset error & $\text{MSE}_\theta^{\mathcal{S}}$ & MSE over torso, arm, and head joints. \\
& Precision gate & $\sigma_{\text{precision}}$ & Soft gate on the reset reward while the object remains at the goal. \\
\bottomrule
\end{tabularx}
\end{table*}

\paragraph{Observation.}
The agent receives proprioception plus SIE, IGT, and HOP features (Table~\ref{tab:observations}). Proprioception is identical across tasks: normalised joint positions, velocities, and torques; joint position and torque command errors; centre-of-mass position and acceleration; and planar ZMP,
\begin{equation}
    p_x = x_{\mathrm{com}} - \frac{z_{\mathrm{com}}\,\ddot{x}_{\mathrm{com}}}{\ddot{z}_{\mathrm{com}} + g},
\qquad
p_y = y_{\mathrm{com}} - \frac{z_{\mathrm{com}}\,\ddot{y}_{\mathrm{com}}}{\ddot{z}_{\mathrm{com}} + g},
\label{eq:zmp}
\end{equation}
with $g = 9.81\,\text{m/s}^2$. Including the ZMP and terminating if it leaves the support polygon discourages unstable whole-body configurations adopted only to accelerate task completion.

Robot and object keypoint states are expressed relative to the robot base, with SIE and object-to-goal edges. Robot keypoints carry Cartesian position, velocity, and a 6D continuous rotation. Object keypoints carry Cartesian position and velocity. Goal keypoints carry Cartesian position ($N_g = N_o$). Each SIE (and object-to-goal) edge contributes a distance vector, scalar distance, relative velocity, and normalised scalar distance. Observation size therefore scales with keypoint count.

\paragraph{Actions.}
AIREC has 35 DoF (head 3, torso 3, dual arms 14, hands 12, omnidirectional base 3). The policy outputs a joint-position delta $\mathbf{a} \in \mathbb{R}^{35}$. We slew-limit the raw delta to $\dot\theta_\text{slew}=1$\,rad/s on revolute joints, then apply an exponential moving average with $\alpha = 0.1$. The result is integrated to a target joint position and tracked by a low-level PD controller, with commanded torque clipped to the actuator limit and joint velocity clipped to $50\%$ of the hardware limit in simulation.

\section{Simulation details}
\label{app:sim}

\paragraph{Scene modelling.}
The bed and wheelchair are modelled to match their real-world counterparts (Figure~\ref{fig:heroshot}). The bed comprises three prisms: a rigid base, a compliant mattress, and a compliant pillow. The wheelchair matches a standard Japanese model with a compact frame ($40$\,cm seat width, versus the $46$\,cm Western standard), so insertion tolerance is tight. Placement into a wheelchair is harder than into an ordinary chair because armrests, side bars, and footrests restrict the viable placement space. For open-loop physical replay, evaluation episodes begin with the robot entirely behind the human, rather than with the mannequin pre-aligned with the robot's hands as during training.

\paragraph{Initialisation, randomisation, and termination.}
In \emph{Ball}, the robot starts $0.3$\,m from the surface with outstretched arms. In the \emph{Human} environments, it starts $0.6$\,m away in a lifting configuration. We randomise robot base pose, object pose, object joints (when applicable), object mass, and goal pose. Episodes terminate when (i) the robot base tilts beyond stability thresholds, (ii) the manipulated object falls below a height threshold, (iii) the object exceeds a distance from the origin, or (iv) a numerical instability is detected.

\begin{figure*}[t]
    \centering
\includegraphics[width=0.99\linewidth]{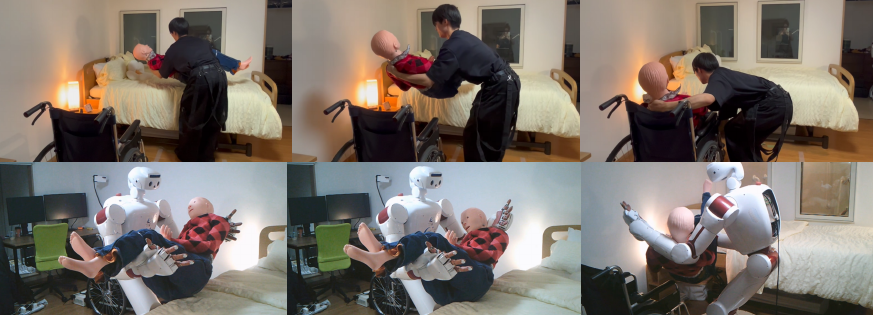}
    \caption{Pathways for imitation learning: mimicking a human demonstration (top), or teleoperating the robot (bottom). An expert teleoperator failed to lift the mannequin robustly with a high-fidelity puppet. The mannequin slipped under its own weight faster than the operator could compensate.}
    \label{fig:placeholder}
\end{figure*}

\section{RL training details}
\label{app:train}

We use clipped PPO~\cite{schulman_proximal_2017}. Parallelism is 4096 environments for \emph{Ball} and 2048 for both \emph{Human} settings. Observations and values are scaled by running normalisers. The policy and critic are MLPs with hidden layers $[512, 256, 128]$ and ELU activations. The policy outputs a Gaussian with a $\tanh$ mean head and a state-dependent $\log\sigma$ head. We use KL-adaptive learning-rate scheduling. Table~\ref{tab:hyperparams} lists the shared PPO recipe; within each environment, our method and the RL baseline use identical optimiser settings. Episode length is $10$\,s on \emph{Ball} and \emph{Human (near)}, and $15$\,s on \emph{Human (far)}. On \emph{Human (far)}, we used a more conservative schedule (learning rate $3\times10^{-4}$, ratio clip $0.15$, $5$ learning epochs) for training stability; the change was applied equally to both methods. With the default schedule, the RL baseline still plateaued.

\begin{table}[h!]
\centering
\caption{PPO and training hyperparameters.}
\label{tab:hyperparams}
\small
\begin{tabular}{lc}
\toprule
\textbf{Parameter} & \textbf{Value} \\
\midrule
Discount $\gamma$ & 0.99 \\
GAE $\lambda$ & 0.95 \\
Learning rate & $5\times10^{-4}$ \\
Ratio clip & 0.2 \\
Value clip & 0.2 \\
Entropy coefficient & 0.001 \\
Value loss scale & 1.0 \\
Learning epochs & 8 \\
Mini-batches & 8 \\
Rollout horizon & 16 \\
Grad norm clip & 1.0 \\
\bottomrule
\end{tabular}
\end{table}

\section{Teleoperation and imitation attempt}
\label{app:imitation}

Imitation offers two routes: mimicking a human caregiver, or teleoperating the robot. We argue that copying human caregiver technique is a limited path for physical assistance, since the technique is constrained by human mobility. We attempted the task via teleoperation with a high-fidelity robotic puppet, but an expert demonstrator failed to lift the mannequin robustly. Under the mannequin's weight, it kept slipping, faster than the operator could react (Figure~\ref{fig:placeholder}).


\end{document}